\documentclass{article}

\usepackage[preprint]{neurips_2025}

\usepackage{placeins} 

\usepackage{graphicx}
\usepackage{subcaption}

\usepackage[utf8]{inputenc} 
\usepackage[T1]{fontenc}    
\usepackage{hyperref}       
\usepackage{url}            
\usepackage{booktabs}       
\usepackage{amsfonts}       
\usepackage{nicefrac}       
\usepackage{microtype}      
\usepackage{xcolor}         
\usepackage{amsfonts}       
\usepackage{amsmath}        
\usepackage{graphicx}

\usepackage{siunitx}
\usepackage{multirow}
\usepackage[font=small,labelfont=bf]{caption}

\title{Image Reconstruction as a Tool for Feature Analysis}

\author{%
  Eduard Allakhverdov \\
  AIRI\\
  Moscow, Russia\\
  MIPT \\
  Dolgoprudny, Russia\\
  \texttt{allakhverdov@2a2i.org} \\
  \And
  Dmitrii Tarasov\\
  AIRI\\
  Moscow, Russia\\
  \texttt{d.tarasov@airi.net} \\
  \And
  Elizaveta Goncharova\\
  AIRI\\
  Moscow, Russia\\
  \texttt{goncharova@airi.net} \\
  \AND
  Andrey Kuznetsov\\
  AIRI\\
  Moscow, Russia\\
  \texttt{kuznetsov@airi.net}
}

\begin{document}

\maketitle

\begin{abstract}
  Vision encoders are increasingly used in modern applications, from vision-only models to multimodal systems such as vision-language models. Despite their remarkable success, it remains unclear how these architectures represent features internally. Here, we propose a novel approach for interpreting vision features via image reconstruction. We compare two related model families, SigLIP and SigLIP2, which differ only in their training objective, and show that encoders pre-trained on image-based tasks retain significantly more image information than those trained on non-image tasks such as contrastive learning. We further apply our method to a range of vision encoders, ranking them by the informativeness of their feature representations. Finally, we demonstrate that manipulating the feature space yields predictable changes in reconstructed images, revealing that orthogonal rotations — rather than spatial transformations — control color encoding. Our approach can be applied to any vision encoder, shedding light on the inner structure of its feature space. The code and model weights to reproduce the experiments are available in GitHub.
\end{abstract}

\section{Introduction}

In recent years, vision encoders have become a crucial component of machine-learning pipelines serving as generalized image representations. They serve as specialized modules for solving computer-vision tasks, aggregating information for image generators (image priors), and powering modern vision–language models. The primary objective of these encoders is to capture all the semantic and object-level information in an image and transfer this knowledge to downstream tasks.

Although experiments can show which model performs best on a specific downstream task, important questions remain: How do hidden representations encode features? What is the relationship between these representations and the original images? Can we manipulate an image by influencing its features – and vice versa?

Answering these questions sheds new light on the interpretability of vision encoders and gives us confidence when choosing and working with their feature outputs. This, in turn, allows for more informed decisions about which encoder is best suited to a particular downstream application.

In this work, we introduce a novel reconstruction-based method to evaluate the quality of features extracted from input images. Our approach reconstructs images from hidden representations, revealing exactly what types of information are preserved within a model’s internal layers.

Our contributions are as follows:
\begin{enumerate}
    \item A novel interpretability metric: We propose an approach for comparing image-feature interpretability based on the visual quality of reconstructed images.
    \item Analysis of training objectives: We perform a detailed comparison between two models with identical architectures – SigLIP \citep{SigLIP} and SigLIP2 \citep{SigLIP2} – whose main difference is their training objective. We demonstrate that SigLIP2 produces significantly higher-fidelity reconstructions than SigLIP.
    \item Models study: We investigate multiple vision-encoder designs --- varying in image resolution, training objectives, and hidden dimensions --- to understand how design choice and peculiarities in model training affect the amount of information each layer retains.
    \item Feature-space transformations: We reveal that applying rotations in the embedding space corresponds to interpretable color transformations in the reconstructed images, pointing toward novel, semantically grounded editing operations.
    \item We release the code to reproduce the results of the experiments available via GitHub repo\footnote{\url{https://fusionbrainlab.github.io/feature_analysis/}}.
\end{enumerate}

\section{Related work}

Recent research has proposed various approaches to interpret the internal representations of vision encoders such as ViT~\cite{dosovitskiy2021an}, CLIP \cite{clip}, DINO \cite{dino}, SAM \cite{kirillov2023segany}, and others. These methods can be grouped into three major categories based on their interpretability strategy.

\subsection{Attention and concept emergence analysis}


Many works focus on analyzing activations to determine which neurons or layers encode specific image concepts—ranging from simple elements like text and colors to more complex structures such as objects and semantics. These approaches typically inspect self-attention maps or neuron activations to reveal what concepts are learned at different stages of transformer-based vision models. For example, \citet{caron2021emerging} demonstrated that attention heads in self-supervised ViTs can function as unsupervised object detectors, effectively segmenting semantically meaningful regions. \citet{dorszewski2025} performed neuron labeling across ViT layers and found a clear progression from low-level features (e.g., edges, color) to high-level concepts (e.g., object parts, categories). More recently, \citet{darcet2024vision} introduced register tokens to eliminate pathological attention patterns and restore clear, interpretable attention in large ViTs. Together, these studies highlight how semantic and spatial regularities can naturally emerge in transformer attention mechanisms --- especially under self-supervised or architecture --- aware training regimes.

\subsection{Representational similarity and probing tasks}

Representational similarity analysis and probing tasks offer large-scale insights into feature spaces. A widely used technique is Centered Kernel Alignment (CKA) \cite{kornblith2019}, which quantifies the similarity between representations from two layers (or two models) on the same inputs --- remaining invariant to orthogonal transformations and isotropic scaling. By computing CKA across all layer pairs, researchers visualize a model's internal structure as a heatmap. \citet{raghu2021} applied CKA to compare ViT and ResNet, discovering that ViT representations remain highly homogeneous across layers, whereas ResNet exhibits a more hierarchical progression with strong locality among adjacent layers. Beyond comparing architectures, CKA has been used to study different training paradigms: supervised versus self-supervised ResNets, or CLIP's image encoder versus a supervised ViT. These analyses consistently find that supervised models concentrate class-discriminative information in deeper layers, while self-supervised models distribute various attributes throughout the network. \citet{kornblith2019} further showed that supervised networks of different architectures converge to remarkably aligned final representations on ImageNet, despite diverging early-layer features.

\subsection{Latent space manipulation}

Latent manipulation techniques operate directly on internal feature vectors --- such as StyleSpace \cite{zongze2021} or CLIP \cite{radford2021learning} embeddings --- to generate counterfactual or modified visual outputs. For instance, StylEx optimizes GAN latents to identify minimal, independent changes in features that affect classifier decisions (e.g., altering the background or object shape)~\cite{lang2021styleex}. \citet{kazemi2024} performed directional edits in CLIP's embedding space to test compositionality and memorization, enabling ``cat astronaut'' image synthesis. Work on CLIP inversion also demonstrates that certain semantic directions correspond to recognizable visual attributes, which can be amplified or suppressed. By enabling fine-grained causal analysis of visual attributes, these methods facilitate the creation of controlled visual counterfactuals and deepen our understanding of model reliance on particular features.

Prior works have explored inversion-based interpretability for convolutional encoders. \citet{Mahendran2016} proposed feature inversion by iteratively optimizing input pixels so that a CNN's activations match those of a target image, revealing which patterns each layer encodes, but their gradient‐descent approach is too slow for real‐time analysis and is tailored to convolutional architectures. \citet{Dosovitskiy2016} advanced this research direction by training a feed‐forward decoder to reconstruct images from CNN features in a single pass, yielding much faster and higher‐fidelity inversion; however, their focus remained on convolutional networks, and they did not examine how different pretraining objectives affect invertibility or the semantic geometry of the feature space. In contrast, our method delivers high‐fidelity, real‐time reconstruction for transformer‐based encoders, enables a direct comparison of contrastive versus multimodal training objectives, and --- by learning orthogonal operators in feature space --- maps latent transformations to precise pixel‐level edits, thus providing a novel tool for probing the structure of modern vision–language embeddings.


\section*{Method}\label{sec:method}

The idea behind the method of vision encoder feature interpretability through reconstruction stems from the following observation. All image encoders represent an input image as a collection of numerical values that, in principle, encode a large part of the information about the image. If we can recover the original image from this internal representation, then the quality of the reconstruction provides a direct measure of how much information the encoder's features preserve.

As an example, let us consider the ViT in a CLIP-base model. An input of size \(224\times224\times3\) pixels is mapped to a feature tensor of shape \(14\times14\times768\), i.e.\ 150\,528 scalar values. By learning an inverse mapping from that tensor back to pixel space, we can potentially reconstruct the original image. The closer the reconstruction is to the true image, the more we know that the encoder's features capture detailed, informative representations.

Once the reconstructor is in place, we can manipulate the feature tensor directly in latent space and observe how those changes manifest in the reconstructed image. In our experiments, we apply controlled orthogonal transformation in feature space that produce systematic color shifts in the output, demonstrating that specific rotations in the encoder's latent space correspond to interpretable visual transformations.

In Section \ref{sec:pipeline}, we provide a detailed description of the introduced method.

\section{Reconstructor pipeline}
\label{sec:pipeline}

\subsection{Reconstructor architecture}

The proposed reconstructor module takes the tensor of the image features as input and provides the initial image reconstruction from this feature tensor. 

Let \(i \in \mathbb{R}^{H \times W \times 3}\) denote an input image, where $H$, $W$ are the width and height of the input image, and $3$ is the number of input channels.  A feature extractor
\[
\mathrm{E} : \mathbb{R}^{H \times W \times C} \;\longrightarrow\; \mathbb{R}^{h \times w \times c}
\]
maps \(i\) to a latent feature tensor \(f = \mathrm{E}(i)\) ($h$ and $w$ are patch size, and $c$ is a hidden dimension).  We then define a reconstruction network
\[
R_{\theta} : \mathbb{R}^{h \times w \times c} \;\longrightarrow\; \mathbb{R}^{H \times W \times C},
\]
parameterized by \(\theta\), whose goal is to approximate the inverse of \(\mathrm{E}\).  In the ideal case, we can reconstruct the same initial image from the latent space $R_{\theta}\bigl(f\bigr) = i$; however, in practice, we obtain 
$\hat{i} = R_{\theta}\bigl(f\bigr),$ where \(\hat{i}\) is the best reconstruction of \(i\) under a chosen loss, for example, mean squared error.

We train \(R_{\theta}\) by minimizing the reconstruction loss $\mathcal{L}_{\mathrm{rec}}$ over the dataset \(\mathcal{D}\).
\[
\mathcal{L}_{\mathrm{rec}} = \mathbb{E}_{i, f \sim \mathcal{D}}\bigl\|\,i - R_{\theta}(f)\bigr\|_2^2
\]

\begin{figure}
    \centering
    \includegraphics[width=1.0\textwidth]{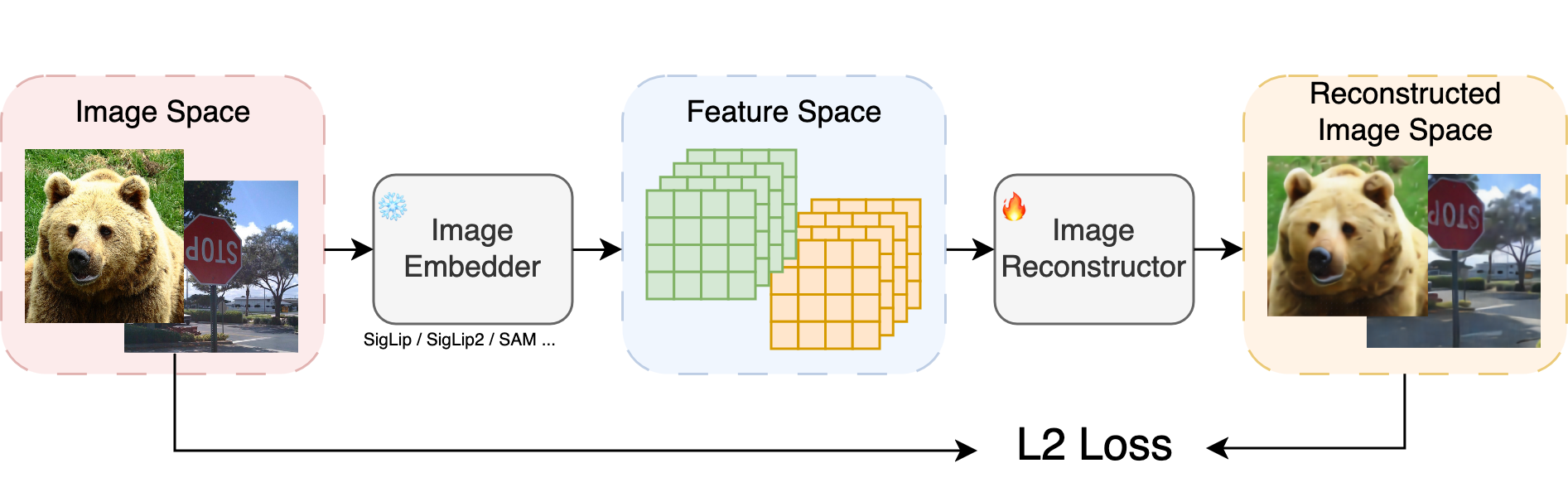}
    \caption{A frozen vision model generates image embeddings, which are then processed by a reconstructor model that learns to approximate image reconstruction.}
    

    \label{fig:image-reconstructor-arch}
\end{figure}

Image encoder is freezed during reconstructor training. Figure~\ref{fig:image-reconstructor-arch} visualize reconstructor training pipeline.

The reconstructor model \(R_{\theta}\) consists of four transformer blocks—each including multi-head self-attention and a feed-forward sublayer—followed by upsampling layers interleaved with residual blocks to restore the spatial resolution from \(h \times w\) back to \(H \times W\). 


\subsection{Reconstructor training}

The reconstructor $R_{\theta}$ is trained in a supervised manner. Let $\{(i_j, f_j)\}_{j=1}^N$ be a dataset of image–feature pairs, where $i_j\in\mathbb{R}^{H\times W\times 3}$ and $f_j = \mathrm{E}(i_j)\in\mathbb{R}^{h\times w\times c}$ ($f$ is calculated as the output of the last hidden state of \(E\)). We optimize $R_{\theta}$ to predict $i_j$ from $f_j$ by minimizing the $\ell_2$ reconstruction loss:
\[
\mathcal{L}_{\mathrm{rec}}
= \frac{1}{N} \sum_{j=1}^N \bigl\lVert\,i_j - R_{\theta}(f_j)\bigr\rVert_2^2.
\]

To improve robustness, we apply channel-wise normalization to each spatial feature vector:
\[
\tilde f_i^{u,v}
= \frac{f_i^{u,v}}{\lVert f_i^{u,v}\rVert_2}
\quad\text{for all }(u,v)\in\{1,\dots,h\}\times\{1,\dots,w\}.
\]
Although this operation discards the original vector norms --- removing magnitude cues --- it suppresses norm-related outliers common in CLIP-style features.

\subsubsection*{Dataset and Feature Extraction}\label{sec:dataset}

To train the image reconstructor \(R_\theta\), we assemble a dataset drawn COCO corpus (CC-BY 4.0 license) \citep{microsoft_coco}. Specifically, we randomly sample 115,000 images for training and reserve an additional 4,000 images for validation.  

For each image \(i_j\) in our dataset, we first compute its feature tensor  
\[
f_j \;=\; E(i_j)\;\in\;\mathbb{R}^{h\times w\times c},
\]  
where \(E\) denotes the frozen vision encoder under study. The reconstructor is then trained to minimize the reconstruction error between the original image \(i_j\) and its estimate  
\[
\hat{i}_j \;=\; R_\theta(f_j).
\]  

Throughout training, we optimize \(\theta\) using an \(\ell_2\) reconstruction loss.


\section{Feature transformation}
\label{sec:transform}

Leveraging our reconstruction-based interpretability pipeline, we can not only recover the input image \(i\) from its feature tensor \(f = E(i)\), but also probe how explicit manipulations in latent space translate back into pixel-space edits. In this section we (1) formulate the hypothesis that certain feature-space operators correspond to well-defined image-space operators, and (2) 
validate it via experiments with color swaps, channel suppression, and colorization. 

\subsection{Hypothesis formulation}

Using the proposed reconstructor pipeline, we are able to recover an image \(i\) from its feature tensor \(f\!=\!\mathrm{E}(i)\). We hypothesize that applying some transformation in latent feature space may correspond to some modifications of the image in pixel space. We formalize the hypothesis relating feature‐space transformations to image‐space transformations as follows.

Let $i \;\in\;\mathbb{R}^{H\times W\times 3}, \quad f \;=\;\mathrm{E}(i)\;\in\;\mathbb{R}^{h\times w\times c}.$

Suppose there exists
\[
A_f\colon \mathbb{R}^{h\times w\times c}\;\to\;\mathbb{R}^{h\times w\times c}, 
\quad
f^* = A_f(f),
\]
\[
A_i\colon \mathbb{R}^{H\times W\times 3}\;\to\;\mathbb{R}^{H\times W\times 3},
\quad
i^* = A_i(i),
\]

where $A_f$ is the transformation in the feature space, and $A_i$ is the corresponding transformation in the image space. Then, $f^*$ corresponds to $i^*$, i.e.\ applying $A_f$ in feature space should correspond to applying $A_i$ in image space.

To test this hypothesis, we compute
\[
f^* = A_f\bigl(\mathrm{E}(i)\bigr),
\quad
\hat{i} = R_{\theta}(f^*).
\]

In other words, transforming \(f\) by \(A_f\) should yield a reconstructed image \(\hat{i}\) that visually matches the pixel‐space transformation \(A_i(i)\).

\subsection{Experimental validation via color manipulations}

\subsubsection{Color swap through reflection}\label{subsec:color-swap}

We evaluate the hypothesis of feature-image space connection through the task of color swap. Let \(A_i\) be an operator that swaps the red and blue channels of \(i\). Following the claim introduced in the previous section, we can find a corresponding \(A_f\) in the latent space such that
\[
R_{\theta}\bigl(A_f(f)\bigr)
\;\approx\;
A_i(i).
\]

\begin{figure}[h!]
    \centering
    \includegraphics[width=1.0\textwidth]{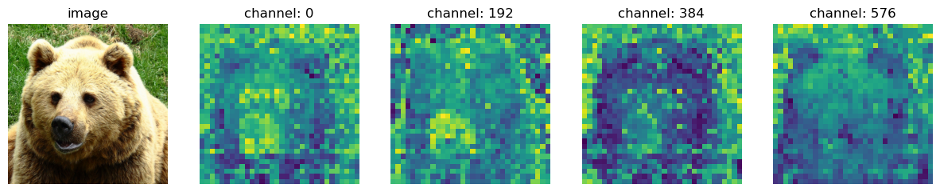}
    \caption{Original image (left) and spatial activation maps for four selected channels of its encoded feature tensor $f$, illustrating how individual feature channels capture coherent image structures.}
    \label{fig:channel-example}
\end{figure}

Let us, first, look into the channels of the feature tensor $f$ for the random image in Figure~\ref{fig:channel-example}.

We can observe for any random image that each channel slice of the reconstructed tensor visually reproduces the original image. If we apply an image-space operator $A_i$
that swaps the red and blue channels of $i$, the reconstructed channels still mirror the modified image. Hence, the corresponding feature-space operator $A_f$ must preserve this per-channel replication property. A natural choice is an orthogonal transformation applied independently to each spatial token. In addition, just as swapping the red and blue channels twice leaves the image unchanged, applying $A_f$ twice must recover the original features. This reversibility suggests that $A_f$ is not only orthogonal but also self-inverse, mirroring the symmetry of the image-space operation.

Formally, let
\[
f = \mathrm{E}(i), 
\quad
i^* = A_i(i),
\quad
f^* = A_f(f).
\]
We posit that
\[
f^*_{k,\ell,\,:} \;=\; Q\,f_{k,\ell,\,:}, 
\quad
Q \in \mathbb{R}^{c\times c}, 
\quad
Q^\top Q = I_{c},
\quad
QQ = I_{c},
\]
for every spatial location \((k,\ell)\).  Under this hypothesis,
\[
\mathrm{E}(i^*) = A_f\bigl(f\bigr),
\quad
R\bigl(\mathrm{E}(i^*)\bigr) = R\bigl(A_f(f)\bigr).
\]

In our experiments, to estimate \(Q\), we collect a dataset of image pairs \(\{(i_j,\,i_j^*)\}_{j=1}^N\) and compute feature pairs
\[
(f_j,\,f_j^*) = \bigl(\mathrm{E}(i_j),\,\mathrm{E}(i_j^*)\bigr),
\quad
f_j, f_j^*\in\mathbb{R}^{h\times w\times c}.
\]
For each spatial index \((k,\ell)\) and image \(j\), we form the paired vectors
\[
\bigl(f_j[k,\ell,:],\,f_j^*[k,\ell,:]\bigr).
\]

\begin{figure}[t!]
    \centering
    \begin{subfigure}[b]{0.47\textwidth}
        \includegraphics[width=\linewidth]{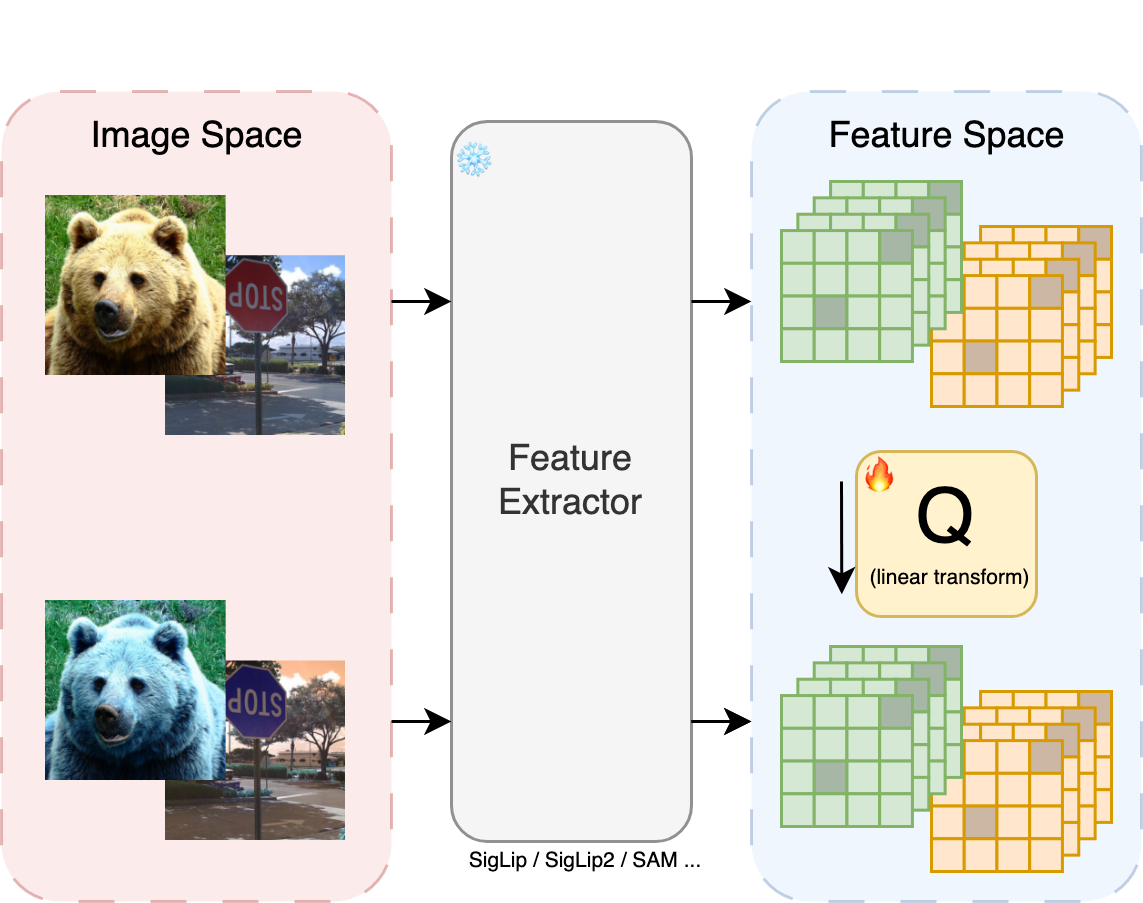}
        \caption{$Q$ Matrix Calculation}
        \label{fig:q_matrix_train}
    \end{subfigure}
    \hspace{0.025\textwidth}%
    \vrule%
    \hspace{0.025\textwidth}%
    \begin{subfigure}[b]{0.40\textwidth}
        \includegraphics[width=\linewidth]{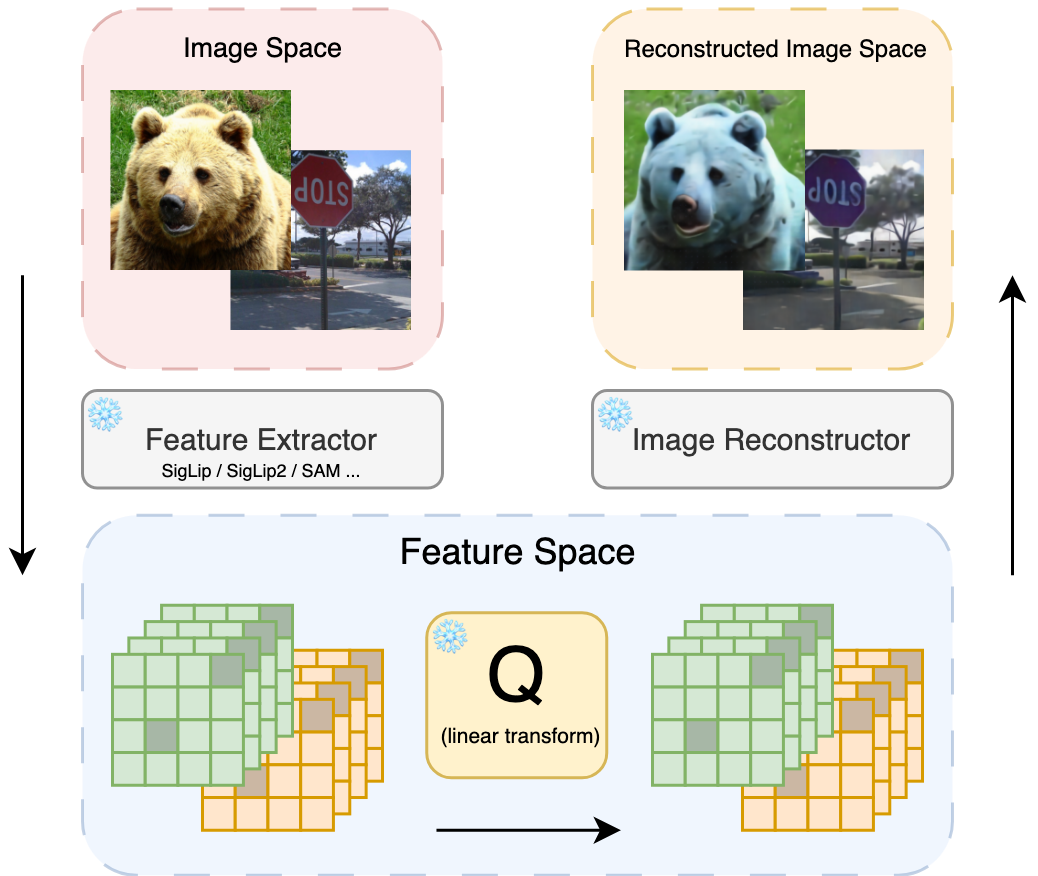}
        \caption{$Q$ Matrix Application}
        \label{fig:q_matrix_train}
    \end{subfigure}
    \hfill
    \caption{Scheme for computing and subsequently applying the matrix \(Q\) in feature space.}
    \label{fig:color-swap-arch}
\end{figure}

We solve the orthogonal Procrustes problem to obtain the matrix \(Q\) that optimally maps \(\{f_j[k,\ell,:]\}\) to \(\{f_j^*[k,\ell,:]\}\) in a least-squares sense. Projecting this matrix onto the subspace of self-conjugated orthogonal operators highlights its self-reversibility property. The computation and subsequent application of \(Q\) are illustrated in Figure~\ref{fig:color-swap-arch}.


\subsubsection{Channel Suppression}\label{subsec:channel-suppression}

To further investigate the connection between feature and image spaces, we examine the operation of channel suppression. Let \( A_i \) denote an operator that attenuates the blue channel of image \( i \) by a factor \( \alpha \in (0, 1) \), with \( A_f \) representing its corresponding operator in the latent space. Following the methodology outlined previously, we construct a dataset \( \left(f_j[k,\ell,:],\,f_j^*[k,\ell,:]\right) \) and use it to learn the linear operator \( A_f \).

We next examine the theoretical properties of \( A_f \). Define \( P_i \) as the operator that zeroes out the blue channel of the image \(i\). Observe that \( A_i^n \;\longrightarrow\; P_i\) as  \( n \to \infty \), and \( P_i \) is a projection operator satisfying \( P_i^2 = P_i \).


Considering the eigendecomposition of \( A_f \), we have:
\[
A_f \mathbf{v} = \lambda \mathbf{v}, \quad \text{and consequently} \quad A_f^n \mathbf{v} = \lambda^n \mathbf{v}.
\]
Since \( A_i^n \) asymptotically approaches a projection operator, its limit \( P_f = \lim_{n \to \infty} A_f^n \) must also possess the properties of a projection operator. This implies that the eigenvalues of \( P_f \) must be either 0 or 1. Therefore, \( \lambda^n \) must converge to either 0 or 1. Therefore, the eigenvalues of \( A_f \) must satisfy either:
\begin{itemize}
    \item \(\lambda = 1\), or
    \item \(\lambda\) is a complex values with magnitude strictly less than 1.
\end{itemize}

These properties are empirically confirmed in Section~\ref{subsec:channel_suppression}.





\subsubsection{Colorization}\label{subsec:colorization}

The experiments described in Sections~\ref{subsec:color-swap} and \ref{subsec:channel-suppression} demonstrate that simple pixel-space transformations correspond to linear transformations in feature space. This observation motivates the question of whether such correspondences extend to more complex, semantically informed transformations beyond algorithmic operations like channel manipulation.

To investigate this, we examine the colorization task - transforming grayscale images to their color counterparts. This problem presents fundamentally different challenges from our previous cases:

1. \textbf{Semantic Requirement}: Successful colorization necessitates that the feature space geometry encodes real-world knowledge about plausible color distributions for objects and scenes.

2. \textbf{Non-algorithmic Nature}: Colorization cannot be achieved via simple pixel‐wise operations; it depends on semantic understanding and context rather than deterministic channel manipulations.

If we can identify a linear transformation in feature space that maps grayscale representations to their colorized counterparts, it would suggest that during training the feature-extractor was able to learn such a geometric space in which the semantics of the real world is translated into the language of this space. Such a result would reveal a non-trivial property of learned feature spaces.

Following our established methodology, we construct a dataset of paired feature vectors \(\left(f_j[k,\ell,:], f_j^*[k,\ell,:]\right)\) representing grayscale and color versions of the same images. We then learn a linear operator \(A_f\) mapping between these representations. The experimental evaluation of this colorization‐based mapping is presented in Section~\ref{subsec:colorization}.

\section{Experiments} \label{sec:experiments}

\subsection{Experimental setup}

\paragraph{Model choice}

To evaluate the insights provided by our reconstruction-based interpretability method, we conduct the main experiments using two vision‐encoder backbones: the SigLIP and SigLIP2 at four input resolutions, \(H \times W = 224\times224\), \(256\times256\), \(384\times384\), and \(512\times512\). These resolutions correspond to output feature tensor shapes of
\[
(h, w, c) \;=\; (14,14,768),\;(16,16,768),\;(24,24,768),\;\text{and}\;(32,32,768),
\]
respectively. SigLIP models are variant of CLIP that uses a sigmoid-based loss.

The motivation for choosing this series of models lies in the similarity of the training and architectural details, whereas the only difference is in the training objectives. 

Both models are pretrained on the WebLI dataset \cite{caron2024webli}: SigLIP uses the English‐only subset, while SigLIP2 employs a multilingual corpus. Although these subsets are not identical, they share the same data source and overall distribution. The key distinction lies in their pretraining objectives: SigLIP is trained solely with a contrastive loss, whereas SigLIP2 additionally incorporates image‐captioning, self‐distillation, and masked‐prediction tasks. We also performed additional experiments on other well-known vision encoders, the experimental results are provided in Appendix \ref{sec:appendix}

\begin{figure}[t!]
    \centering
    \begin{subfigure}[b]{0.29\textwidth}
        \includegraphics[width=\linewidth]{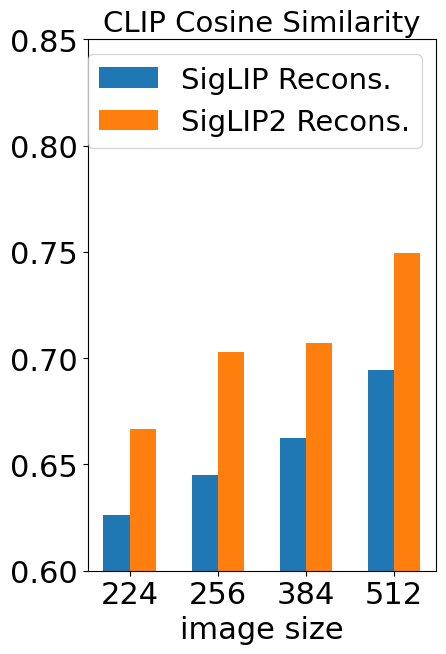}
        \caption{CLIP-Score}
        \label{fig:SigLIP12_clip_reconstruction_metrics}
    \end{subfigure}
    \begin{subfigure}[b]{0.29\textwidth}
        \includegraphics[width=\linewidth]{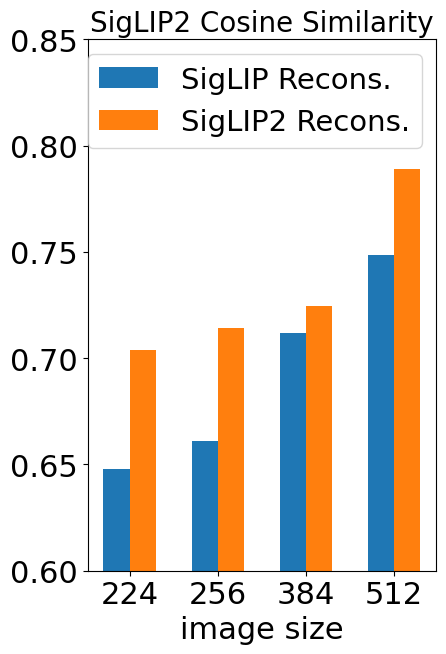}
        \caption{SigLIP2-Score}
        \label{fig:SigLIP12_SigLIP_reconstruction_metrics}
    \end{subfigure}
    \hfill
    \begin{subfigure}[b]{0.365\textwidth}
        \includegraphics[width=\linewidth]{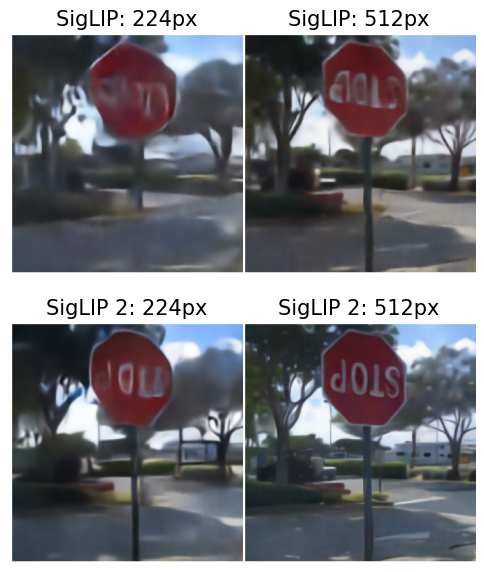}
        \caption{Reconstruction Samples}
        \label{fig:SigLIP12_reconstruction_samples}
    \end{subfigure}
    
    \caption{Comparison of SigLIP and SigLIP2 reconstruction performance and visual results.}
    \label{fig:SigLIP12_combined}
\end{figure}

For our evaluation, we reconstruct images from the COCO‐val split and measure the fidelity of their latent representations. Let \(i\) be an original validation image and \(\hat{i} = R_{\theta}(\mathrm{E}(i))\) its reconstruction (cf. Section~\ref{sec:method}). 
Following \cite{clipscore} we compute cosine‐similarity scores between the encoder's embeddings of \(i\) and \(\hat{i}\):
\[
\mathrm{sim}_{\mathrm{orig},\mathrm{rec}}
= \cos\bigl(S(i),\,S(\hat{i})\bigr),
\]
where \(S\) is instantiated as either the CLIP or SigLIP2 encoder. Figure~\ref{fig:SigLIP12_combined} reports these similarity scores alongside representative reconstruction examples from the COCO‐val set. We employ Wilcoxon signed-rank and bootstrap tests to demonstrate SigLIP2's statistically significant superiority over SigLIP in reconstruction quality across all evaluated resolutions ($p < 0.01$ for all comparisons), with detailed results presented in Table~\ref{tab:reconstruction_compact}.

\begin{table}[h!]
\centering
\caption{Comparison of reconstruction quality between SigLIP and SigLIP2 evaluated using google-SigLIP2-large-patch16-256 and openai-clip-vit-large-patch14. Reported are p-values from Wilcoxon signed-rank tests and bootstrap tests ($n=1000$ samples, $B=100,000$ resampling iterations). Null hypothesis ($H_0$): equivalent reconstruction quality; alternative hypothesis ($H_1$): SigLIP2 provides superior reconstruction quality}
\label{tab:reconstruction_compact}

\vspace{\baselineskip}

\setlength{\tabcolsep}{12pt}
\renewcommand{\arraystretch}{1.1}
\begin{tabular}{@{}lS[table-format=1.2e-1]S[table-format=1.2e-1]S[table-format=1.2e-1]S[table-format=1.2e-1]@{}}
\toprule
\multicolumn{1}{c}{Resolution} & 
\multicolumn{2}{c}{CLIP-ViT} & 
\multicolumn{2}{c}{SigLIP2} \\
\cmidrule(lr){2-3} \cmidrule(lr){4-5}
 & {Wilcoxon} & {Bootstrap} & {Wilcoxon} & {Bootstrap} \\
\midrule
$224 \times 224$ px & 1.5e-79 & <1e-300 & 1.4e-113 & <1e-300 \\
$256 \times 256$ px & 2.3e-123 & <1e-300 & 1.2e-103 & <1e-300\\
$384 \times 384$ px & 3.9e-82 & <1e-300 & 2.3e-08 & 2.2e-04 \\
$512 \times 512$ px & 2.0e-111 & <1e-300 & 3.8e-64 & <1e-300 \\
\bottomrule
\end{tabular}
\end{table}

\subsection{Color swap}\label{subsec:color_swap}
To evaluate whether swapping image color channels induces a systematic shift in the learned feature space, we randomly sample 1024 images from the COCO validation set and train $Q$ using the procedure described in Section~\ref{sec:transform}. Next, we use the trained reconstructor $R_\theta$ to invert modified latents back to the RGB domain. Finally, we evaluate the entire pipeline on the remaining COCO‐val images unseen during the training of both $Q$ and $R_\theta$. Example reconstructions for SigLIP2 model are shown in Figure~\ref{fig:color-swap-mini}. We also provide additional examples in Appendix (Figure \ref{fig:color-swap}).

\begin{figure}
    \centering
    \includegraphics[width=1.0\linewidth]{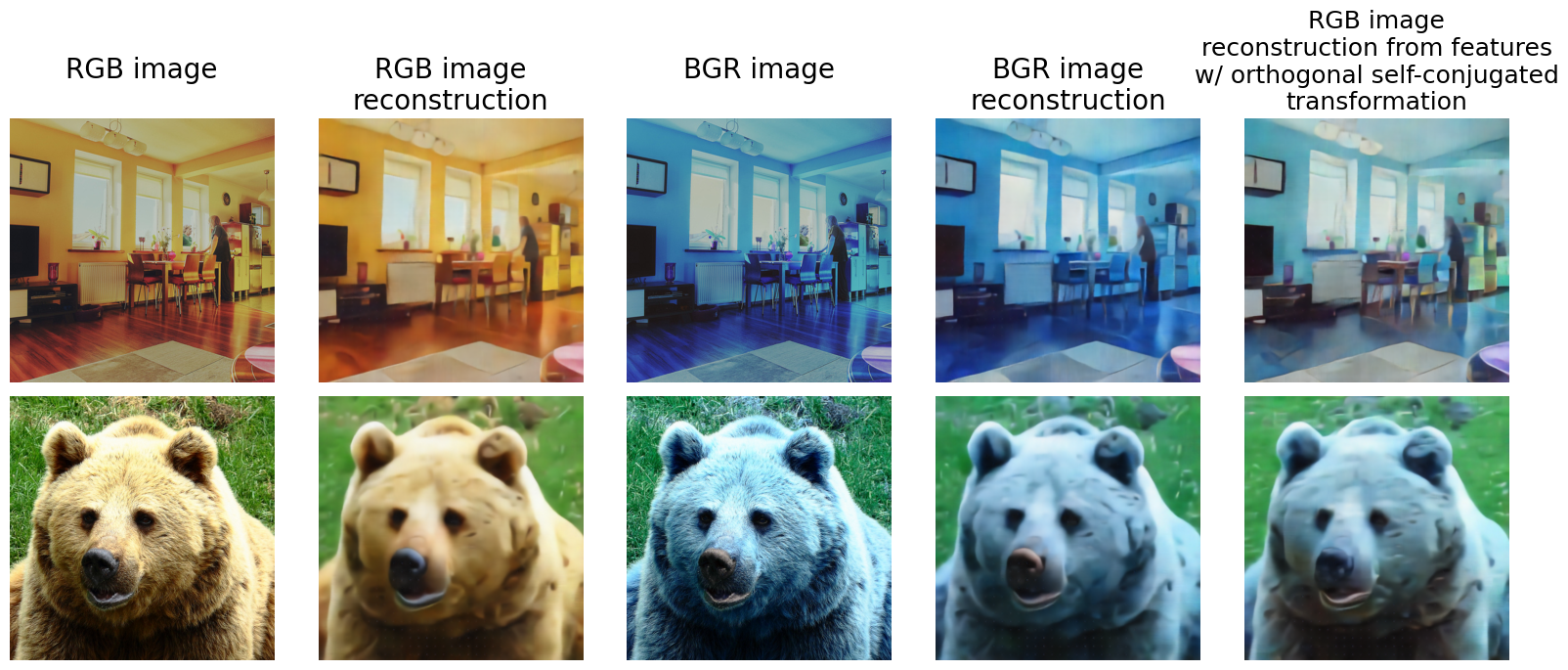}
    \caption{Color‐swap via orthogonal rotations in SigLIP2 feature space. Each row presents: (1) the original image, (2) its reconstruction from encoder features, (3) the image after swapping red and blue channels in pixel space, (4) the reconstruction of the pixel‐swapped image, and (5) the reconstruction obtained by applying the corresponding orthogonal self-conjugated channel‐swap directly in feature space. The near‐identical results in columns 4 and 5 confirm that simple rotations in latent space induce coherent, interpretable color edits in the reconstructed images.}
    \label{fig:color-swap-mini}
\end{figure}

We observe that reconstructing from the transformed latent representations reproduces the channel‐swapped image. Interestingly, this demonstrates that a color permutation in pixel space corresponds to a orthogonal transformation in the feature space.

\subsection*{Color swap operator properties ablations}\label{subsec:color_swap_properties_ablation}

As detailed in Section~\ref{sec:transform}, the feature-space operator $A_f$ is constrained to be both \textbf{orthogonal} and \textbf{self-conjugated} to maintain consistency with the properties of the pixel-space operator $A_i$. In this subsection, we conduct an ablation study examining the effects of these $A_f$ properties --- specifically its orthogonality and self-conjugacy.


\begin{figure}
    \centering
    \includegraphics[width=1.0\linewidth]{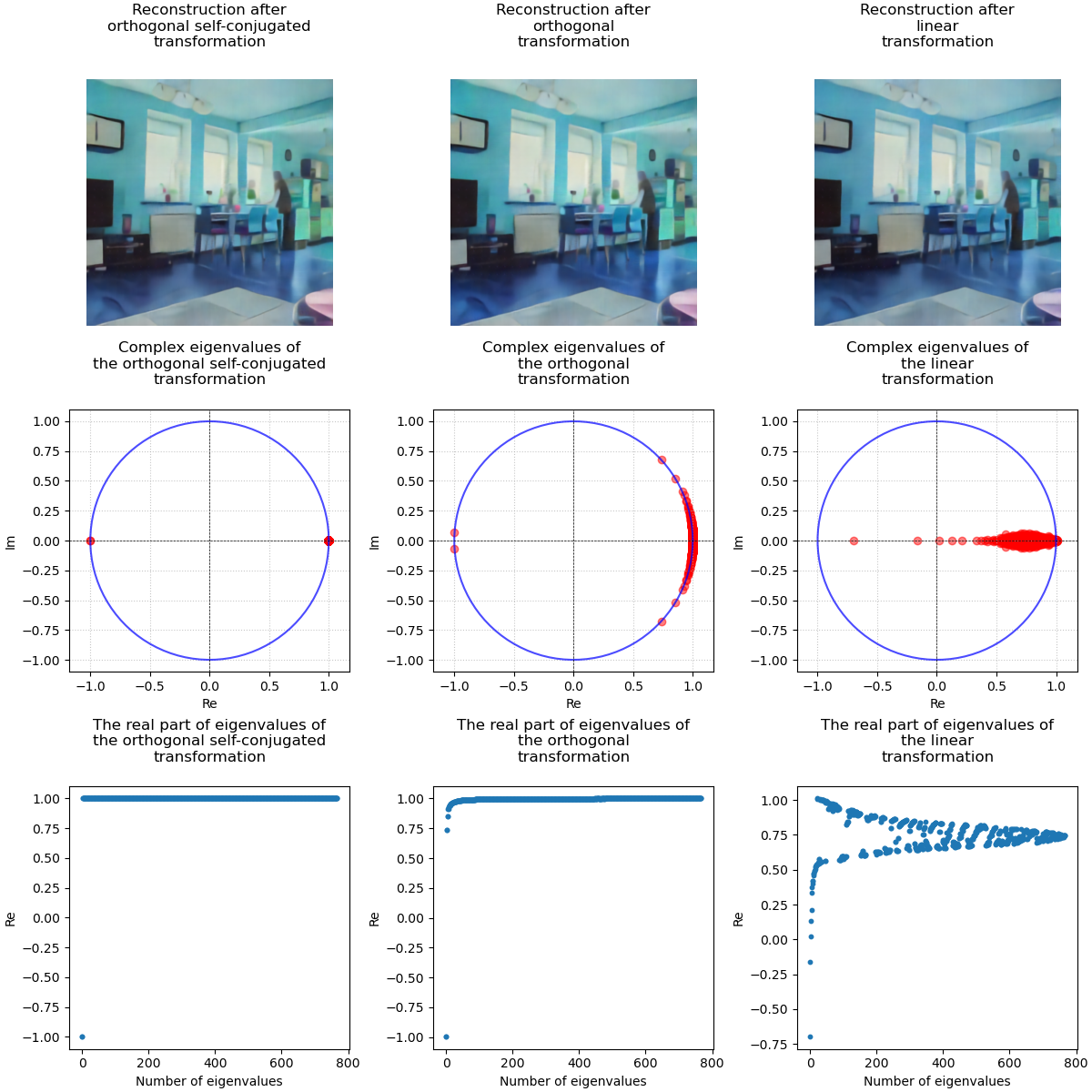}
    \caption{ Color swap operator properties ablation. \textbf{Rows:} (1) sample for transformed image reconstruction, (2) $A_f$ operator eigenvalues visualization, (3) values of real part of $A_f$ eigenvalues. \textbf{Columns:} (1) self-conjugated and orthogonal operator constraints, (2) orthogonal only constraint (3) no constraints }
    \label{fig:color-swap-eigen-values}
\end{figure}

We trained three different operators:
\begin{enumerate}
    \item Orthogonal self-conjugated --- as a Procrustes solution with a long-range projection of the operator onto the space of self-conjugated operators.
    \item Orthogonal --- as a Procrustes solution.
    \item Linear --- as a regression problem. (Note that this solution cannot be directly used with the reconstructor, as it fails to preserve vector norms. Since the reconstructor was trained exclusively on normalized vectors, we first normalize the resulting outputs before feeding them to the reconstructor.).
\end{enumerate}

As shown in Figure~\ref{fig:color-swap-eigen-values}, the eigenvalues of all operators cluster along the real axis, indicating they primarily represent either eigenvector preservation (near +1) or inversion (near -1). While small deviations from these ideal values exist - revealing noise in the feature space - these perturbations remain relatively weak. Consequently, the feature space geometry largely preserves the properties expected from the pixel-space channel permutation operator.

\subsection{Channel suppression}\label{subsec:channel_suppression}
\begin{figure}
    \centering
    \includegraphics[width=1.0\linewidth]{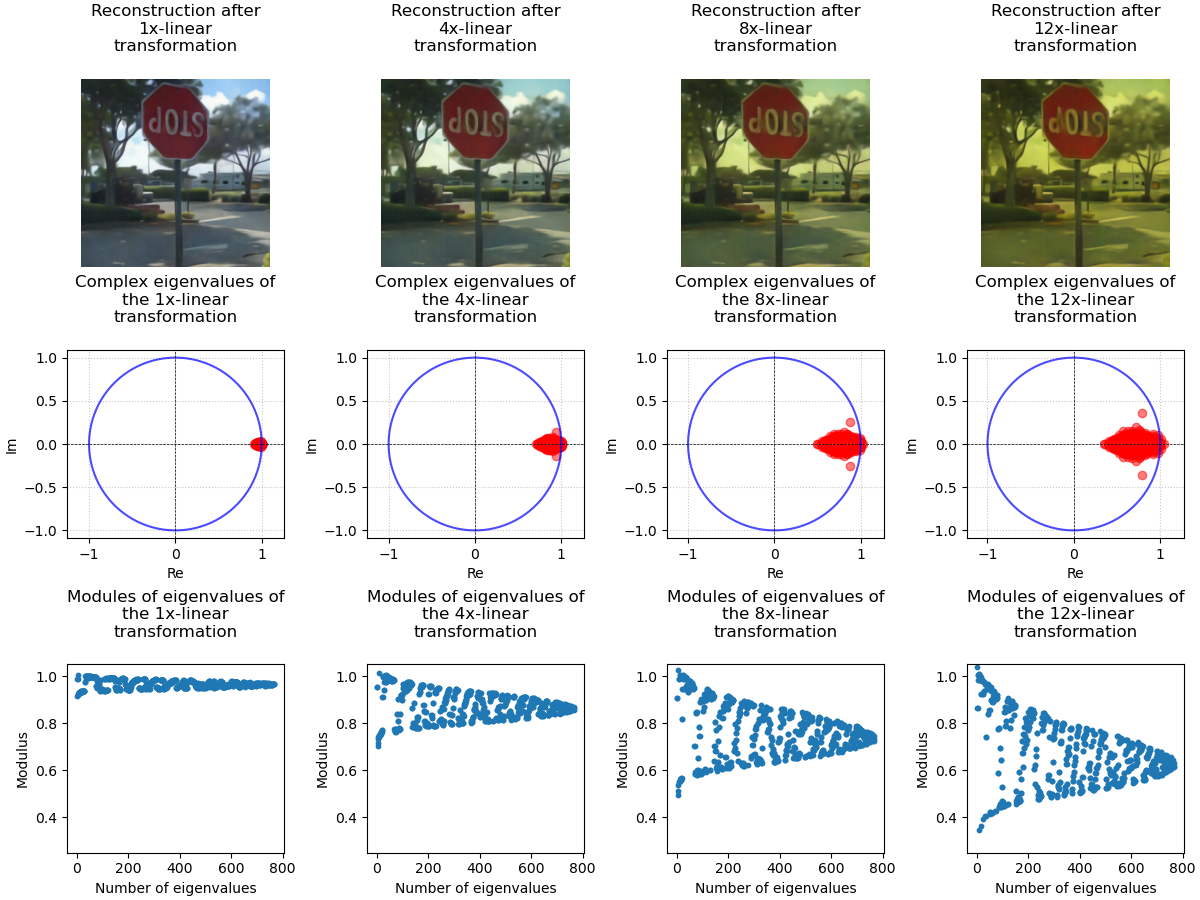}
    \caption{Ablation of operator properties for blue‐channel suppression with coefficient \(\alpha = 0.9\). \textbf{Rows:} (1) Sample for transformed image reconstruction, (2) $A_f$ operator eigenvalues visualization, (3) Values of real part of $A_f$ eigenvalues. \textbf{Columns:} (1) Single application of the operator , (2) Quadruple operator application, (3) Eightfold operator application, (4) Twelvefold operator application. As predicted, the eigenvalues are either equal to 1 or inside the circle.}
\label{fig:b_suppression_all_eigen_values}
\end{figure}

To evaluate the hypothesis proposed in Section~\ref{subsec:channel-suppression} we use the same method as in Section~\ref{subsec:color_swap}, i.e., train a linear operator on pairs of features for the initial image and the image with suppressed blue channel, then apply this operator on new images and check that the reconstruction visually coincides with the corresponding transformation in the pixel space. (Similar to the linear operator from Section~\ref{subsec:color_swap}, we normalize the features after exiting the linear transformation for the correct operation of the reconstructor.)

In addition, we check whether the eigenvalues of the resulting operator are either equal to $1$ or have norm less than $1$ (see Section~\ref{subsec:channel-suppression}).


As shown in Figure~\ref{fig:b_suppression_all_eigen_values}, the eigenvalues of the operator do obey the predicted property. The same figure shows that the repeated application of the operator behaves similarly to the projector as described in Section~\ref{subsec:channel-suppression}. Additional examples are provided in Appendix (Figure \ref{fig:b-suppression}).

\subsection{Colorization}\label{subsec:colorization}
\begin{figure}
    \centering
    \includegraphics[width=1.0\linewidth]{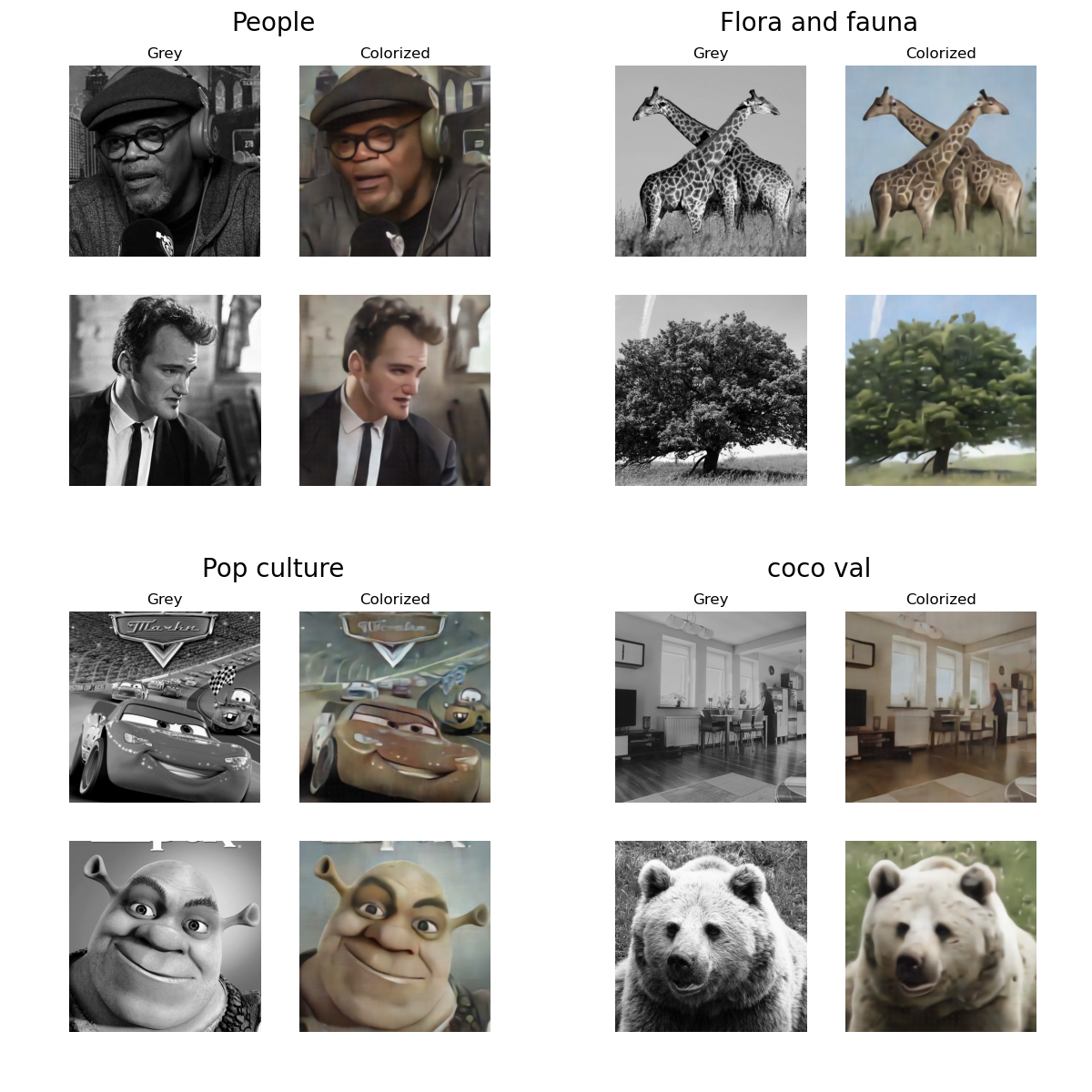}
    \caption{Examples of solving the colorization problem by applying a linear transformation in the feature space.}
    \label{fig:colorized_examples}
\end{figure}

The colorization experiments follows the same methodology as in the previous sections. Specifically, we train a linear operator \(A_f\), and during inference we first apply \(A_f\) to each feature chip and then normalize these chips to ensure the reconstructor functions correctly.

Figure~\ref{fig:colorized_examples} demonstrates the results of such a colorization. We observe that the colorization method performs accurately on objects with inherently unambiguous color distributions. For instance, entities such as human skin, trees, grass, sky regions, and animals with naturally uniform coloration are reliably colorized.

These experiments support our hypothesis that semantically meaningful transformations can be represented linearly in the learned feature space, although the performance boundaries are determined by the consistency of color distributions in the real world. We also provide additional examples in Appendix (Figure \ref{fig:colorized_all_examples}).

\section{Limitations}
\label{sec:limitations}

After training the reconstructor $R_\theta$, we observe that $R_\theta$ serves as an approximate inverse of the feature extractor $\mathrm{E}$. However, an exact inverse cannot be obtained due to the following factors:

\begin{enumerate}
  \item \textbf{Non‐invertibility of $\mathrm{E}$}
  
    In general, $\mathrm{E}$ is not required to be bijective. As evidenced by the experiments reported in Section \ref{sec:experiments}, different input resolutions yield reconstructions of varying quality, even though 
    \[
      H\times W\times 3 
      \;=\;(H/16)\times(W/16)\times\bigl(3\times16^2\bigr)
    \]
    remains constant. Higher input resolution makes $\mathrm{E}$ more nearly invertible, but does not guarantee exact recovery.
    
  \item \textbf{Finite training coverage.}  
    We train $R_\theta$ on a limited dataset, so the learned inverse may generalize poorly to regions of feature space that are underrepresented. If manipulated features fall into such poorly covered regions, the reconstructor can produce inconsistent or implausible outputs.
\end{enumerate}

\subsection{Addressing limitations for the interpretablity pipeline}

The first limitation is inherent: for a fixed $\mathrm{E}$, there is an upper bound on the fidelity any reconstructor can achieve. The second --- namely, how well the reconstructor generalizes to regions of feature space not covered by the training set --- remains an open question and merits further investigation in future work.

In the context of the task solved by the proposed paper, we treat the reconstructor as the reflector of the quality of the constructed vision latents, and the amount of information kept in them. Thus, we address the limitations written above in the following manner.

\paragraph{Non-invertibility of E} In our experiments (see Figure \ref{fig:SigLIP12_reconstruction_samples}), we observed that the residual inversion error of the encoder \(E\) manifests as small, localized distortions in the reconstructed images. Consequently, before applying this method to a particular task, one should verify that the magnitude of these distortions is acceptable --- and in practice these minor artifacts do not impede our ability to validate the analyzed feature --- image transformation hypothesis, since they affect all compared reconstructions equally.

\paragraph{Finite training coverage} To address the second limitation about the incomplete coverage of feature-space regions by the reconstructor, we require that the reconstructor and the transformation pair be conditionally independent given the encoder:
  \[
    R_{\theta}\;\perp\!\!\!\perp\;(A_i, A_f)\;\bigl|\;E.
  \]
This means that \(R_{\theta}\) is trained without any knowledge of the specific image‐ or feature‐space operators \(A_i\) and \(A_f\) beyond what it already learns from \(E\).
 
The operators \(A_i\) and \(A_f\) are defined independently of the reconstructor and rely only on information encoded in \(E\).

Ensuring this conditional independence guarantees that our validation of the \(A_f\leftrightarrow A_i\) correspondence is not biased by artifacts introduced during reconstructor training.

We consider the channel swapping experiment for the above limitations in \ref{sec:color-swap-limitation-analysis}.

We note that our current experiments are restricted to ViT‐based architectures. In future work, we plan to extend our reconstruction‐based interpretability framework to other model classes, in particular convolutional neural networks \cite{convnet}.

\section{Conclusion}

In this paper, we propose a novel, reconstruction‐based interpretability framework. By learning an approximate inverse mapping from encoder features back to pixels, our method enables controlled manipulations in feature space, and directly reveals their corresponding effects in image space.  

During experimental evaluation, we applied our reconstruction‐based interpretability framework to a diverse set of ViT‐based vision encoders that differ in training objectives, pretraining datasets, and hidden dimensions. We paid particular attention to the SigLIP/SigLIP2 family, which share identical architectures, parameter counts, and datasets but differ only in their optimization objectives.

Our study demonstrates three fundamental findings:
\begin{enumerate}
  \item \textbf{Resolution and informativeness of the feature space:} For a fixed architecture, the amount of visual information retained in the latent space increases with the spatial resolution of the feature tensor.
  \item \textbf{Role of training objectives:} Features learned using image‐based objectives exhibit richer visual semantics and better preserve structural details.
  \item \textbf{Mathematics of transformations:} Edits in image‐channel space correspond to orthogonal transformations of token embeddings in the latent space.
\end{enumerate}

For future work, we plan to extend our comparison to models with different architectures --- such as modern convolutional neural networks (e.g., ConvNext~\cite{convnet}) --- and to conduct a deeper, layer‐wise analysis of the information encoded at each transformer depth.





{
\small

\bibliographystyle{apalike}
\bibliography{my_ref}

}


\appendix

\section{Appendix} \label{sec:appendix}

\subsection{Reconstruction of the various encoders}

As an ablation study, we train the reconstructor for several of the most well-known vision encoders that are widely used in computer vision applications. Specifically, we select multiple families of ViT-based encoders that vary in parameter count, architectural details, training objectives, and pretraining datasets. In Table~\ref{tab:image_encoders_seq_length}, we present the architectural analysis of the evaluated vision encoders. Tables~\ref{tab:image_model_objectives_text} and~\ref{tab:image_model_objectives_checkmarks} show training peculiarities of the analyzed encoders.


\begin{table}[h!]
\centering
\caption{Comparison of Image Encoders: input resolution, sequence length, parameter count, embedding dimension}
\label{tab:image_encoders_seq_length}
\vspace{\baselineskip}
\begin{tabular}{@{\centering\arraybackslash}p{4cm} r r >{\centering\arraybackslash}p{2cm} >{\centering\arraybackslash}p{1.5cm}@{}}
\toprule
\textbf{Model} & \textbf{Resolution} & \textbf{Sequence dength} & \textbf{\#Params {\scriptsize Vision tower}} & \textbf{Out dimension} \\
\midrule
    \textbf{CLIP} \\

    timm/EVA-02 \cite{Fang2024}           & 224×224 px   & 196 {\scriptsize(14×14)} & 86 M  & 768  \\
    OpenAI CLIP \cite{clip}               & 224×224 px   & 196 {\scriptsize(14×14)} & 86 M  & 768  \\
    LAION CLIP \footnote{https://laion.ai/blog/large-openclip/}                           & 224×224 px   & 196 {\scriptsize(14×14)} & 86 M  & 768  \\
    Facebook MetaCLIP \cite{xu2024metaclip} & 224×224 px & 196 {\scriptsize(14×14)} & 86 M  & 768  \\
    Apple DFN2B-CLIP \cite{fang2023data}  & 224×224 px   & 196 {\scriptsize(14×14)} & 86 M  & 768  \\

    \\ \textbf{SigLIP} \\

    SigLIP \citep{SigLIP}                 & 224×224 px   & 196 {\scriptsize(14×14)} & 93 M  & 768  \\
    SigLIP \citep{SigLIP}                 & 256×256 px   & 256 {\scriptsize(16×16)} & 93 M  & 768  \\
    SigLIP \citep{SigLIP}                 & 384×384 px   & 576 {\scriptsize(24×24)} & 93 M  & 768  \\

    \\ \textbf{SigLIP2} \\

    SigLIP2 \citep{SigLIP2}               & 224×224 px   & 196 {\scriptsize(14×14)} & 93 M  & 768  \\
    SigLIP2 \citep{SigLIP2}               & 256×256 px   & 256 {\scriptsize(16×16)} & 93 M  & 768  \\
    SigLIP2 \citep{SigLIP2}               & 384×384 px   & 576 {\scriptsize(24×24)} & 93 M  & 768  \\
    SigLIP2 \citep{SigLIP2}               & 512×512 px   & 1024 {\scriptsize(32×32)}& 93 M  & 768  \\

    \\ \textbf{SAM} \\

    Facebook SAM \cite{kirillov2023segany} & 1024×1024 px & 4096 {\scriptsize(64×64)} & 90 M & 768  \\

    \\ \textbf{DinoV2} \\

    DinoV2 \citep{dinov2} & 518×518 px & 1369 {\scriptsize(37×37)} & 87M & 768  \\

    \\ \textbf{InternViT} \\

    InternViT-V1.5-300M \cite{gao2024miniinternvl} & 448×448 px & 1024 {\scriptsize(32×32)} & 304 M & 1024 \\
    InternViT-V2.5-300M \cite{chen2025internvl25} & 448×448 px & 1024 {\scriptsize(32×32)} & 304 M & 1024 \\

\bottomrule
\end{tabular}
\end{table}

\begin{table}[h!]
\centering
\caption{Reconstructor Training Hyperparameters}
\label{tab:reconstructor_training_hyperparams}
\vspace{\baselineskip}
\begin{tabular}{ll}
\toprule
\textbf{Hyperparameter} & \textbf{Value} \\
\midrule
Optimizer & Adam \citep{adam} \\
Learning Rate & 3e-4 \\
Learning Rate Scheduler & Cyclic \\
Adam Beta1 & 0.9 \\
Adam Beta2 & 0.999 \\
Batch Size (per device) & 10 \\
Training Epochs & 40 \\
\bottomrule
\end{tabular}
\end{table}

\begin{table}[h!]
\caption{Comparison of Image Encoders: training objective and architecture}
\label{tab:image_model_objectives_text}
\centering
\vspace{\baselineskip}
\begin{tabular}{@{}lp{10cm}}  
\toprule
\textbf{Model} & \textbf{Training Objective} \\
\midrule

    \textbf{CLIP} \\

    timm/EVA-02 & Contrastive image–text alignment. Weights inited from EVA model trained on Masked image modeling: reconstructing CLIP features from masked patches (negative cosine loss) \\
    
    OpenAI CLIP & Contrastive image–text alignment (InfoNCE) \\

    LAION CLIP & Contrastive on LAION-2B (2 B image–text pairs) \\
    
    Facebook MetaCLIP & Contrastive on CommonCrawl 2.5 B data  \\

    Apple DFN2B-CLIP & Contrastive on DFN-2B filtered data \\

    \\ \textbf{SigLIP} \\
    
    SigLIP & Pairwise sigmoid contrastive loss \\
    
    \\ \textbf{SigLIP2} \\
    
    SigLIP2 & Multitask: sigmoid contrastive, captioning, self-distillation, masked modeling  \\

    \\
    \textbf{SAM} \\

    Facebook SAM & Promptable segmentation: predict masks from sparse or dense prompts  \\

    \\ \textbf{DinoV2} \\

    DinoV2  &  Discriminative self-supervised pretraining: self-distillation from Dino \citep{dino} and masked image modeling from iBOT \citep{ibot}  \\

    \\ \textbf{InternViT} \\
    
    InternViT-V1.5-300M & Contrastive pre-training, LLM alignment, distillation \\
    
    InternViT-V2.5-300M & Contrastive pre-training, LLM alignment with progressive scaling, distillation  \\

\bottomrule
\end{tabular}
\end{table}

\begin{table}[h!]
\caption{Comparison of Image Models by Pretraining Objectives}
\label{tab:image_model_objectives_checkmarks}
\centering
\vspace{\baselineskip}
\begin{tabular}{@{}%
  >{\raggedright\arraybackslash}p{2.8cm}  
  >{\centering\arraybackslash}p{1.5cm}  
  >{\centering\arraybackslash}p{1.4cm}  
  >{\centering\arraybackslash}p{1.5cm}  
  >{\centering\arraybackslash}p{1.4cm}  
  >{\centering\arraybackslash}p{1.4cm}    
  >{\centering\arraybackslash}p{1.4cm}    
@{}}
\toprule
\textbf{Model}                                & \textbf{Contrastive} & \textbf{Captioning} & \textbf{Masked Modeling} & \textbf{Segmentation} & \textbf{LLM Alignment} & \textbf{Self-Distillation} \\
\midrule

\textbf{CLIP} \\  
timm/EVA-02                                   & \checkmark          &                      & \checkmark               &                      &                        \\
OpenAI CLIP                                   & \checkmark          &                      &                          &                      &                        \\
LAION CLIP                                    & \checkmark          &                      &                          &                      &                        \\
Facebook MetaCLIP                             & \checkmark          &                      &                          &                      &                        \\
Apple DFN2B-CLIP                              & \checkmark          &                      &                          &                      &                        \\

\\ \textbf{SigLIP} \\  
SigLIP                                        & \checkmark          &                      &                          &                      &                        \\

\\ \textbf{SigLIP2} \\  
SigLIP2                                       & \checkmark          & \checkmark          & \checkmark               &                      &                        \\

\\ \textbf{SAM} \\  
Facebook SAM                                  &                      &                      &                          & \checkmark           &                        \\

\\ \textbf{DinoV2} \\  
DinoV2                                  &                      &                       &  \checkmark                         &            & & \checkmark                        \\

\\ \textbf{InternViT} \\  
InternViT-V1.5-300M                           & \checkmark          &                      &                          &                      & \checkmark             \\
InternViT-V2.5-300M                           & \checkmark          &                      &                          &                      & \checkmark             \\

\bottomrule
\end{tabular}
\end{table}

\begin{figure}[h!]
    \centering
    \includegraphics[width=1.0\textwidth]{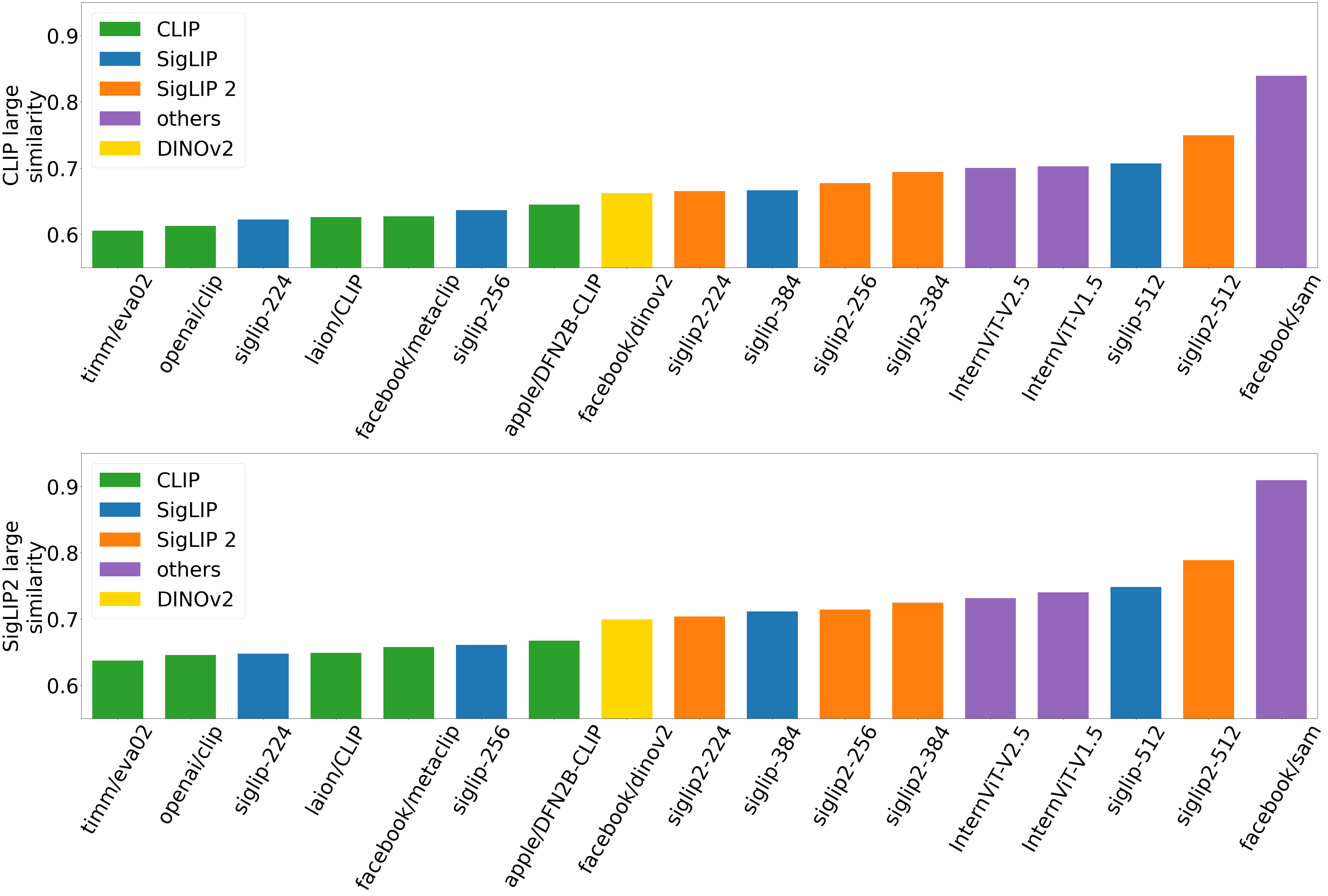}
    \caption{Encoder performance comparison on the COCO val set, showing average CLIP similarity and SigLIP2 similarity between original images and their reconstructions for each vision encoder. Higher bars indicate better alignment of reconstructed images with the originals under each metric.}
    \label{fig:all-encoders-comparison}
\end{figure}

\begin{figure}
    \centering
    \includegraphics[width=1.0\textwidth]{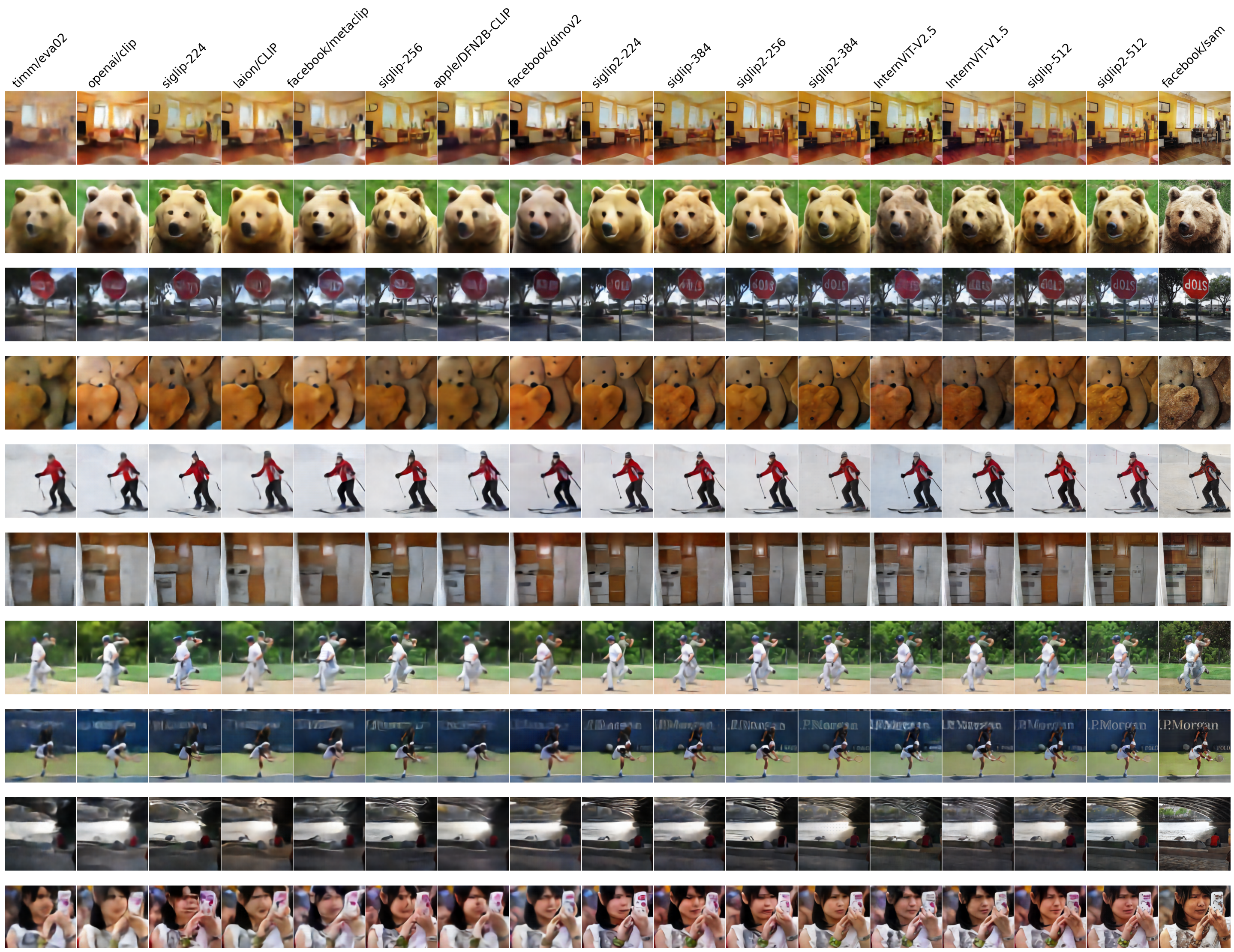}
    \caption{Qualitative comparison of original images and their reconstructions obtained from different vision encoders.}
    \label{fig:all-encoders-images}
\end{figure}

Figure \ref{fig:all-encoders-comparison} presents the CLIP and SigLIP scores for original images and their reconstructions—obtained via the trained reconstructor \(R_\theta\) --- on the COCO validation set across various encoders. Figure \ref{fig:all-encoders-images} provides qualitative examples of original images alongside their reconstructions generated from the latent representations of different vision encoders.

These results align with our main findings from the SigLIP–SigLIP2 comparison using the reconstructor method. We also observe that higher input resolution produces richer latent representations and better reconstructions, as seen in the InternViT-448, SigLIP2-512, and SAM-1024 models. Finally, vision-based pretraining objectives yield superior feature encodings, resulting in higher similarity scores.

\subsection{Color Swap limitation analysis}\label{sec:color-swap-limitation-analysis}

To check the correctness of the result with swapping channels, make sure that:
\begin{enumerate}
    \item The characteristic size of the distortion is smaller than the characteristic size of the object affected by the feature transformation.
    \item  \(R_{\theta}\;\perp\!\!\!\perp\;(A_i, A_f)\;\bigl|\;E\), where
    \(E\) takes an image as input and returns a tensor of features, \(R_{\theta}\) reconstructs the image from the feature tensor, \(A_i\) swaps picture channels in pixel space, \(A_f\) performs orthogonal transformation in the space of features.

\end{enumerate}

In such terms for our experiment we have:
\begin{enumerate}
    \item Swapping channels is an operation that works on the whole image at once, and the size of the image is much larger than the size of the characteristic distortion for any model in our experiments.
    \item \(A_i\) operator changes image channels in places purely algorithmically, \(A_f\) during training relies only on representations obtained with \(E\) and images changed with \(A_i\), and \(R_{\theta}\) relies only on representations obtained with \(E\). The training datasets of \(A_f\) and \(R_{\theta}\) do not overlap.
\end{enumerate}

\begin{figure}
    \centering
    \includegraphics[width=1.0\textwidth]{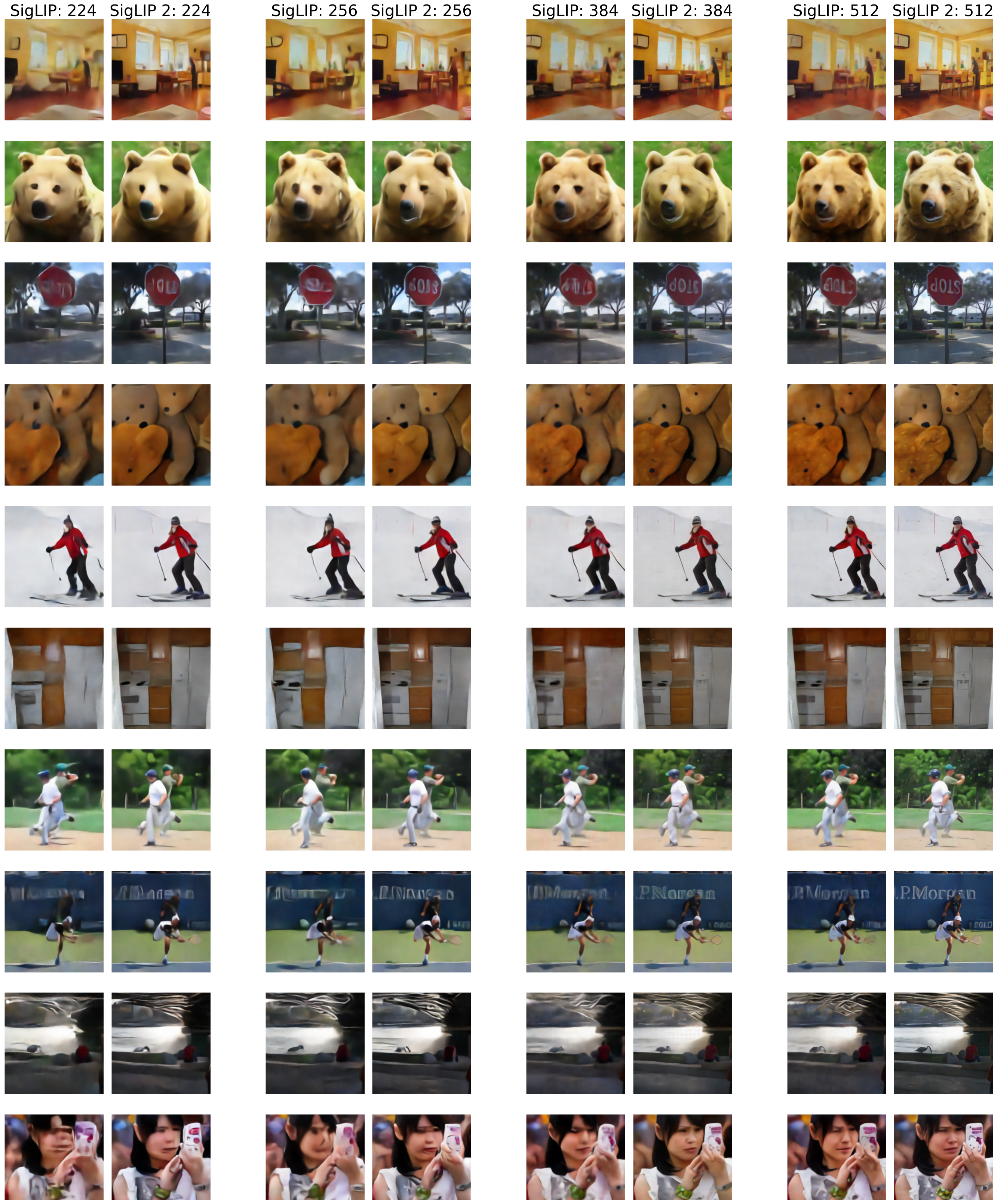}
    \caption{Qualitative analysis of the SigLIP and SigLIP2 reconstruction samples.}
    \label{fig:SigLIP12_reconstruction_sapmles}
\end{figure}



\begin{figure}
    \centering
    \includegraphics[width=0.78\linewidth]{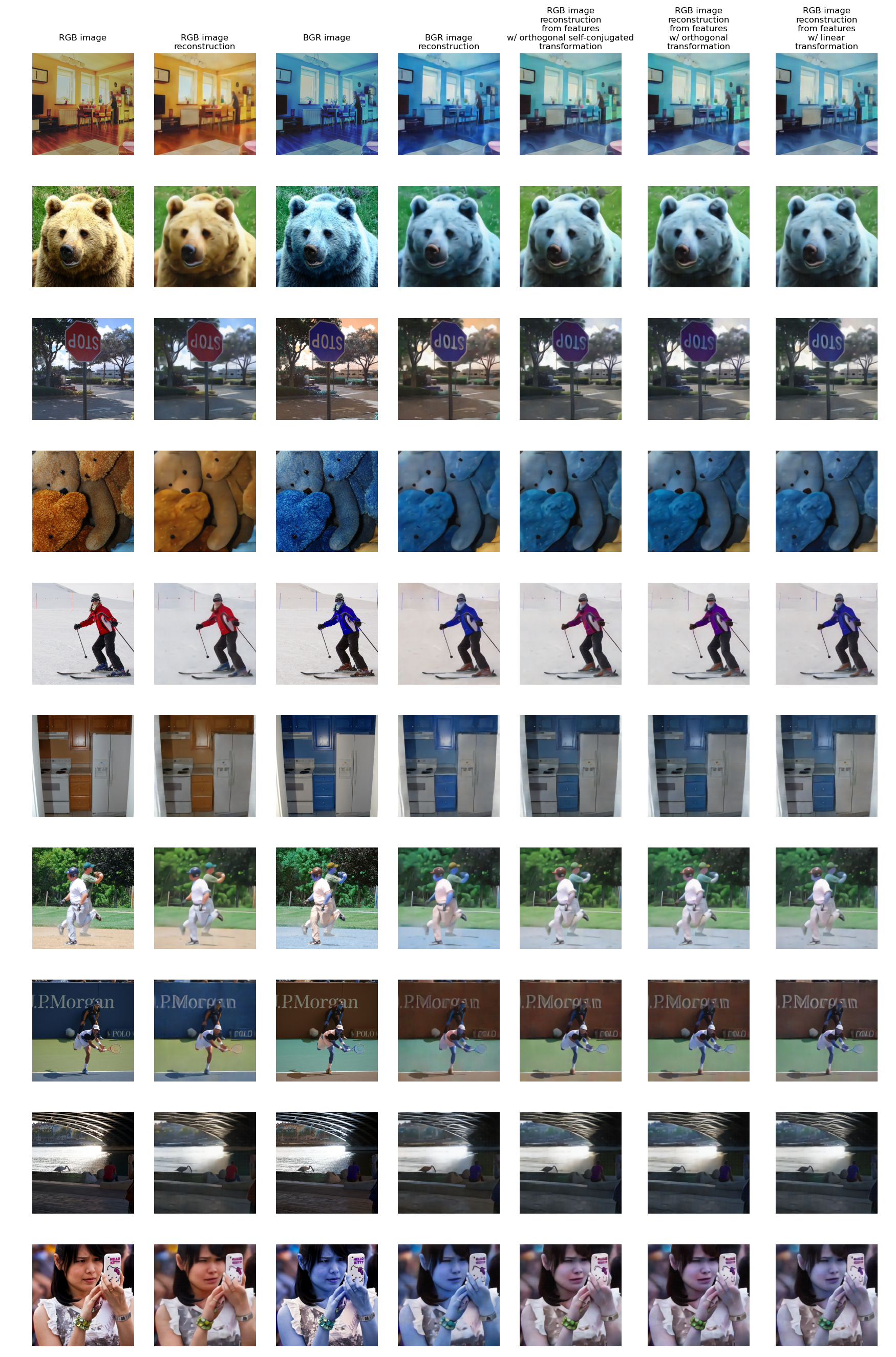}
    \caption{Color-swap via simple transformations in SigLIP2 feature space. Each row presents: (1) the original image, (2) its reconstruction from encoder features, (3) the image after swapping red and blue channels in pixel space, (4) the reconstruction of the pixel-swapped image, (5) the reconstruction obtained by applying the corresponding orthogonal self-conjugated channel-swap directly in feature space, (6) the reconstruction obtained by applying the corresponding orthogonal channel-swap directly in feature space, (7) the reconstruction obtained by applying the corresponding linear channel-swap directly in feature space. The near-identical results in columns 4 and 5, 6, 7 confirm that simple transformations in latent space induce coherent, interpretable color edits in the reconstructed images.}
    \label{fig:color-swap}
\end{figure}

\begin{figure}
    \centering
    \includegraphics[width=0.78\linewidth]{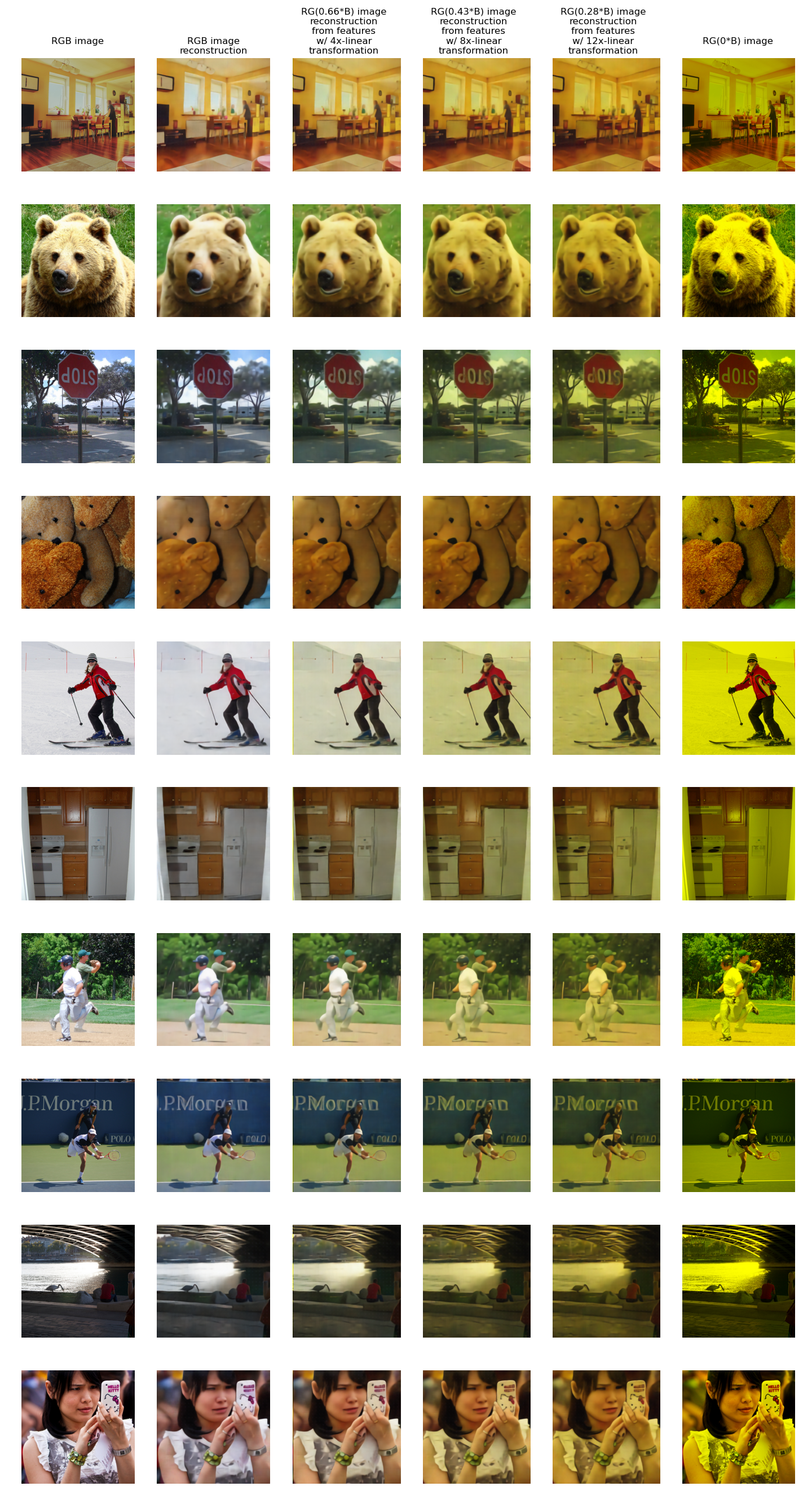}
    \caption{B-channel suppression via linear transformations in SigLIP2 feature space. Each row presents: (1) the original image, (2) its reconstruction from encoder features (3) reconstruction obtained by quadrupling the corresponding linear blue channel suppression operator directly in the feature space, (4) reconstruction obtained by applying the corresponding linear blue channel suppression operator eight times directly in the fisheye space, (5) reconstruction obtained by twelvefold the corresponding linear blue channel suppression operator directly in the feature space, (6) the image after blue channel nulling in pixel space.
    The near-identical results in columns 5 and 6 confirm that simple transformations in latent space induce coherent, 
    interpretable color edits in the reconstructed images.}
    \label{fig:b-suppression}
\end{figure}

\begin{figure}
    \centering
    \includegraphics[width=1\linewidth]{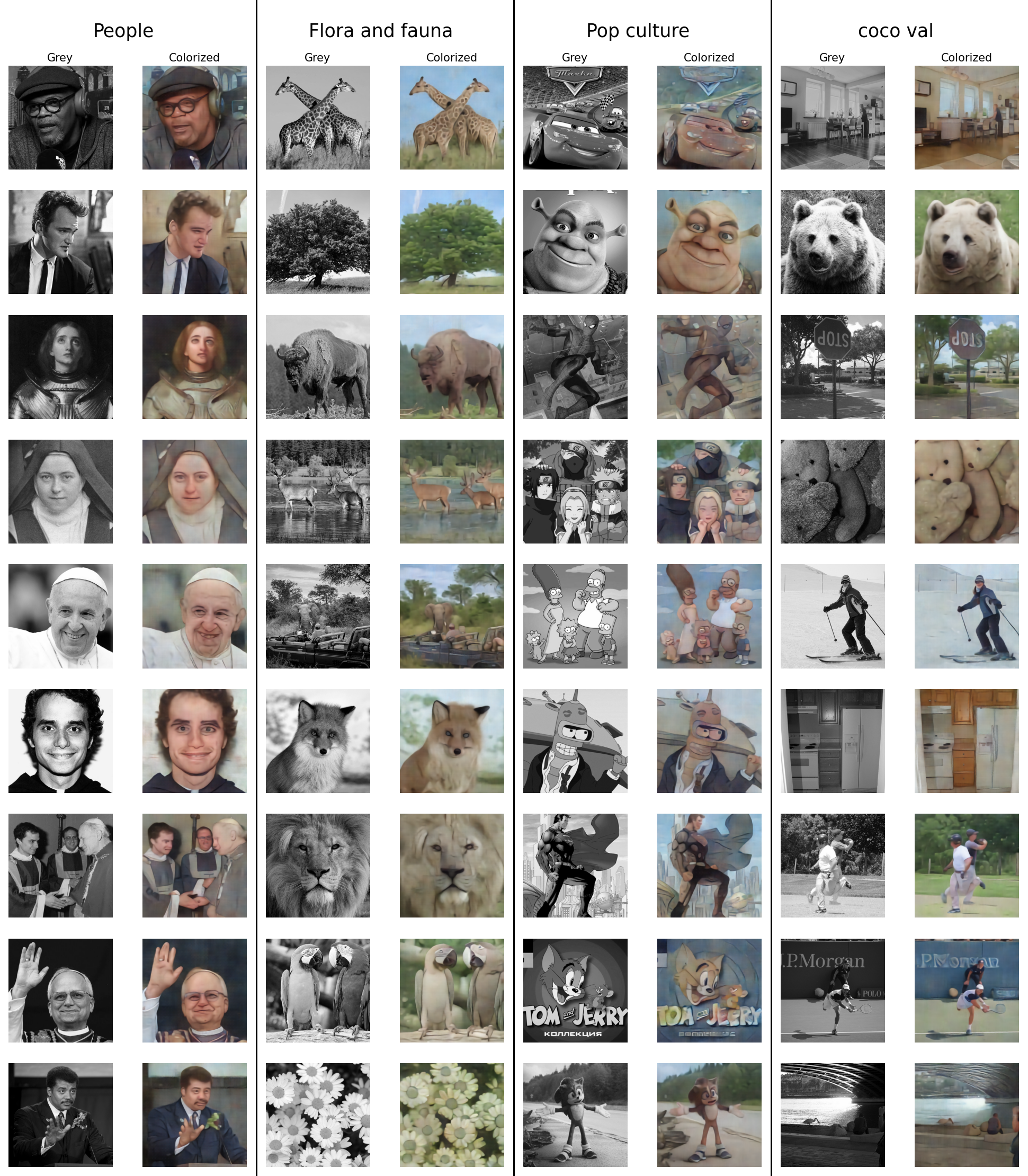}
    \caption{Examples of solving the colorization problem by applying a linear transformation in the feature space.}
    \label{fig:colorized_all_examples}
\end{figure}

\subsection{Qualitative analysis for SigLIP and SigLIP2}

In Figure~\ref{fig:SigLIP12_reconstruction_sapmles} we present the qualitative analysis of the reconstructions obtained for SigLIP and SigLIP2 models across four different resolutions. In Figure~\ref{fig:color-swap} we present the qualitative analysis for Color Swap experiments described at Section~\ref{subsec:color_swap}.

\subsection{Training Details}\label{sec:training_details}


Reconstructors were trained with Adam and a cyclic learning‐rate schedule. Hyperparameters are listed in Table~\ref{tab:reconstructor_training_hyperparams}, and data details in Sec.~\ref{sec:dataset}. Training on 3×A100 (80 GB) GPUs took 6–24 h, depending on resolution.

\clearpage
\FloatBarrier

\end{document}